\documentclass[sigconf]{acmart}
\AtBeginDocument{%
  \providecommand\BibTeX{{%
    \normalfont B\kern-0.5em{\scshape i\kern-0.25em b}\kern-0.8em\TeX}}}

\copyrightyear{2021}
\acmYear{2021}
\setcopyright{acmlicensed}\acmConference[SIGIR '21]{Proceedings of the 44th
International ACM SIGIR Conference on Research and Development in
Information Retrieval}{July 11--15, 2021}{Virtual Event, Canada}
\acmBooktitle{Proceedings of the 44th International ACM SIGIR Conference on
Research and Development in Information Retrieval (SIGIR '21), July 11--15,
2021, Virtual Event, Canada}
\acmPrice{15.00}
\acmDOI{10.1145/3404835.3463011}
\acmISBN{978-1-4503-8037-9/21/07}

\usepackage{multirow}
\usepackage{balance}
\newcommand{\be}{\mathbf{e}}
\newcommand{\bm}{\mathbf{m}}
\newcommand{\bv}{\mathbf{v}}

\newcommand{\bh}{\mathbf{h}}

\newcommand{\bM}{\mathbf{M}}
\newcommand{\eg}{\textit{e.g.}}
\newcommand{\ie}{\textit{i.e.}}

\begin{document}
\fancyhead{} 
\title{Proactive Retrieval-based Chatbots based on Relevant Knowledge and Goals}

\author{Yutao Zhu$^1$, Jian-Yun Nie$^1$, Kun Zhou$^2$, Pan Du$^1$, Hao Jiang$^3$, and Zhicheng Dou$^4$}
\affiliation{%
  $^1$ Université de Montréal, Montréal, Québec, Canada \\
  $^2$ School of Information, Renmin University of China, Beijing, China \\
  $^3$ Huawei Poisson Lab., Hangzhou, Zhejiang, China \\
  $^4$ Gaoling School of Artificial Intelligence, Renmin University of China, Beijing, China
  \country{}
}
\email{yutao.zhu@umontreal.ca,{nie,pandu}@iro.uimontreal.ca,francis\_kun\_zhou@163.com}
\email{jianghao66@huawei.com,dou@ruc.edu.cn}


\renewcommand{\authors}{Yutao Zhu, Jian-Yun Nie, Kun Zhou, Pan Du, Hao Jiang, and Zhicheng Dou}

\begin{abstract}
A proactive dialogue system has the ability to proactively lead the conversation. Different from the general chatbots which only react to the user, proactive dialogue systems can be used to achieve some goals, \eg, to recommend some items to the user. Background knowledge is essential to enable smooth and natural transitions in dialogue. In this paper, we propose a new multi-task learning framework for retrieval-based knowledge-grounded proactive dialogue. To determine the relevant knowledge to be used, we frame knowledge prediction as a complementary task and use explicit signals to supervise its learning. The final response is selected according to the predicted knowledge, the goal to achieve, and the context. Experimental results show that explicit modeling of knowledge prediction and goal selection can greatly improve the final response selection. Our code is available at \url{https://github.com/DaoD/KPN/}.
\end{abstract}




\maketitle

\section{Introduction}
From Microsoft Xiaoice, Apple Siri, to Google Assistant, dialogue systems have been widely applied in our daily life. In general, these systems are designed to make responses in reaction to the user's requirements, such as play music, set a clock, or show the weather forecast. These systems are perceived just as tools by users 
as they only react passively. Users may be bored quickly.
The problem is even more severe in a chit-chat style dialogue system. \textbf{Proactive conversation} offers a solution to this problem~\cite{tang-etal-2019-target,DBLP:conf/acl/WuGZWZLW19,DBLP:conf/iccai/YuanA20}. 
A proactive dialogue system can lead the dialogue proactively to achieve some goals. 
For example, it can drive the conversation to educate the kids about some topics, to comfort a person, to place ads unobtrusively, or to recommend items to users. 
Various application scenarios are emerging, 
yet there has not been a standard definition of what proactive conversation should be. Variations have been observed in the way that a goal is defined: by a sentence~\cite{zhu-etal-2020-scriptwriter} or by some entities that should be covered in the conversation~\cite{DBLP:conf/acl/WuGZWZLW19}, and whether the goal should be generated dynamically~\cite{tang-etal-2019-target} or predefined~\cite{DBLP:conf/iccai/YuanA20}.
We are still in the early stage of exploration in which people test different approaches in different settings. This study is a contribution to this exploration. 

\begin{figure}[!t]
    \centering
    \small
    \setlength{\abovecaptionskip}{0.1cm}
    \setlength{\belowcaptionskip}{0cm}
    \includegraphics[width=\linewidth]{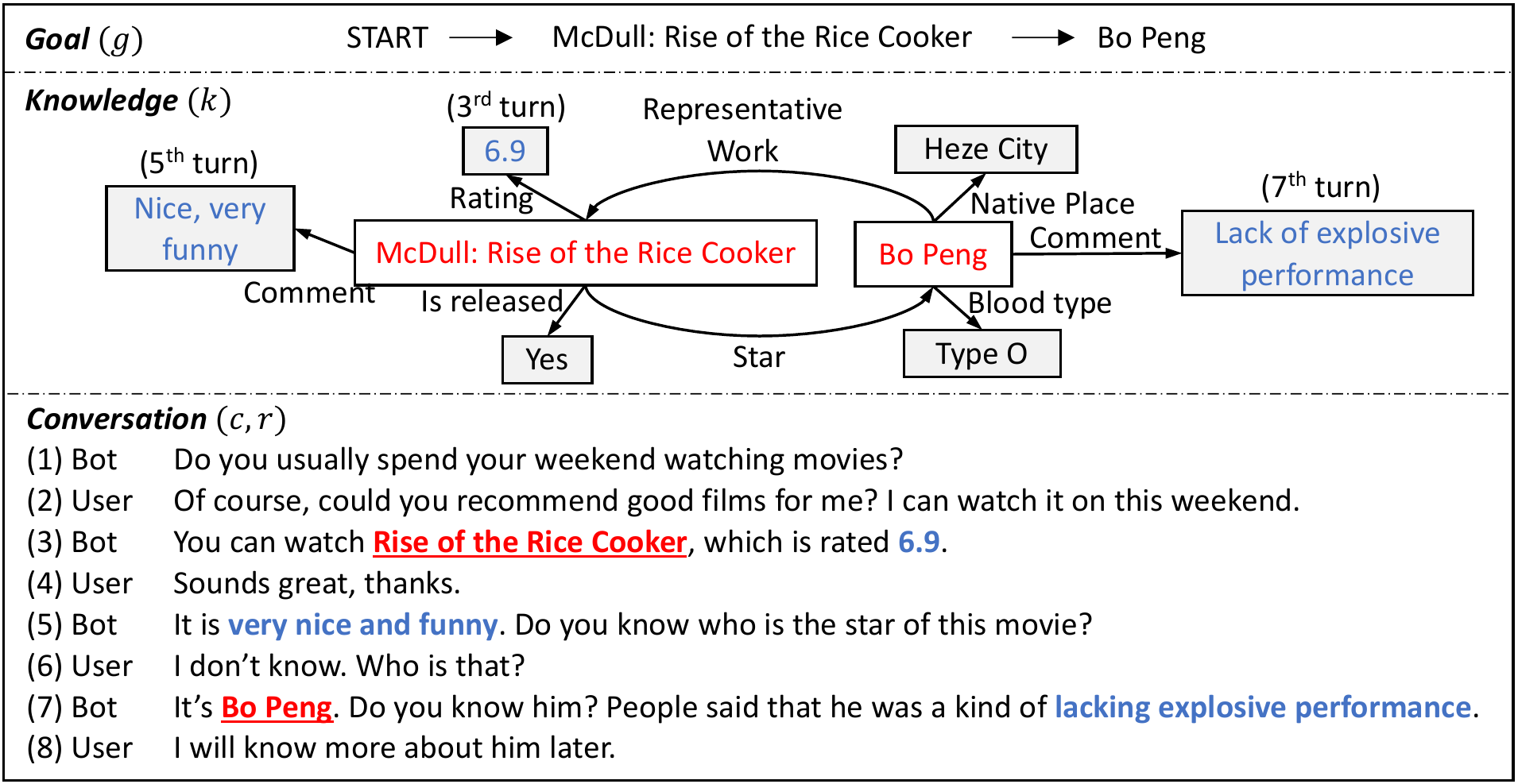}
    \caption{An example of proactive dialogue. The system is asked to exploit the background knowledge to lead the dialogue and accomplish the goal.} 
    \label{fig:example}
    \vspace{-10pt}
\end{figure}

In this work, we follow the setting given by~\citet{DBLP:conf/acl/WuGZWZLW19}, where the goal is specified by a set of entities (topics) and the background knowledge about these entities is provided in a knowledge graph. The goal is to lead the conversation smoothly to mention the required entities. The knowledge graph helps to generate paths of conversation that appear natural.
Despite its simplicity, this setting has many potential applications in practice, in particular in conversational recommendation systems~\cite{DBLP:conf/nips/LiKSMCP18,kang-etal-2019-recommendation,DBLP:conf/aaai/QinYTL20}, where some items can be set in advance for recommendation.
An example of  proactive conversation in the movie domain is shown in Figure~\ref{fig:example}. The goal is defined by two entities (topics): the movie \textit{McDull: Rise of the Rice Cooker} and the star \textit{Bo Peng}. The system is asked to cover both entities during the conversation. By exploiting the knowledge graph, the system aims to naturally transit from one conversation topic to another and eventually fulfill the pre-defined goal. 

In this work, we focus on the retrieval-based method due to its higher fluency.
Although knowledge has been incorporated in some existing approaches to proactive conversation~\cite{DBLP:conf/acl/WuGZWZLW19}, it has been simply embedded in the response selection process~\cite{DBLP:conf/acl/LiuFCRYL18,DBLP:conf/aaai/GhazvininejadBC18,DBLP:conf/ijcai/LianXWPW19}, 
which is optimized globally according to the loss on the final response selection. Although the end-to-end training could be reasonable with a very large amount of training data, in practice, the limited training data may lead to sub-optimal solutions: when a wrong response is selected by the system, it is hard to tell if it is due to a poor knowledge prediction or a bad response selection, thus hard to optimize. 


To tackle this problem, we design an explicit \textit{Knowledge Prediction} (KP) module to select the relevant piece of knowledge to use. This module is combined with a \textit{Response Selection} (RS) module, and both form a multi-task learning framework, called \textbf{Knowledge Prediction Network} (KPN). 
The two tasks are jointly learned. The KP module first tracks the state of goal achievement, \ie, which part of the goal has been achieved,  and then leverages the dialogue context to predict which knowledge should be used in the current turn. The RS module then relies on the selected knowledge to help select the final answer.
Different from the existing methods, we explicitly optimize KP using automatically generated weak-supervision signals to help better learn to predict the relevant knowledge. 
Experimental results show that the explicitly trained KP process can indeed select the most relevant piece of knowledge to use, and this leads to superior performance over the state-of-the-art methods. 

Our main contributions are two-fold:
(1) We propose a multi-task learning framework for knowledge-grounded proactive dialogue, in which the knowledge prediction task is explicitly trained in a weakly supervised manner. 
(2) We show experimentally that our model significantly outperforms the existing methods, demonstrating the great importance of knowledge selection in proactive conversation. 

\section{Knowledge Prediction Network}
\textbf{Problem Formalization} We follow the task definition formulated by~\citet{DBLP:conf/acl/WuGZWZLW19}.
For a dataset $\mathcal{D}$, each sample is represented as $(c,g,k,r,y)$ (as shown in Figure~\ref{fig:example}), where $c=\{u_1,\cdots,u_{L}\}$ represents a conversation context with $\{u_{i}\}_{i=1}^{L}$ as utterances; $g$ represents the goal containing some entities that the dialogue should talk about (\eg, ``Bo Peng''); $k=(k_1,\cdots,k_{M})$ are knowledge triplets where $k_i$ is in form of SPO (Subject, Predicate, Object); $r$ is a response candidate; $y \in \{0,1\}$ is a binary label. The task is to learn a matching model $s(c,g,k,r)$ with $\mathcal{D}$ to measure the suitability of a response candidate $r$. 

\begin{figure}[!t]
    \centering
    \includegraphics[width=.75\linewidth]{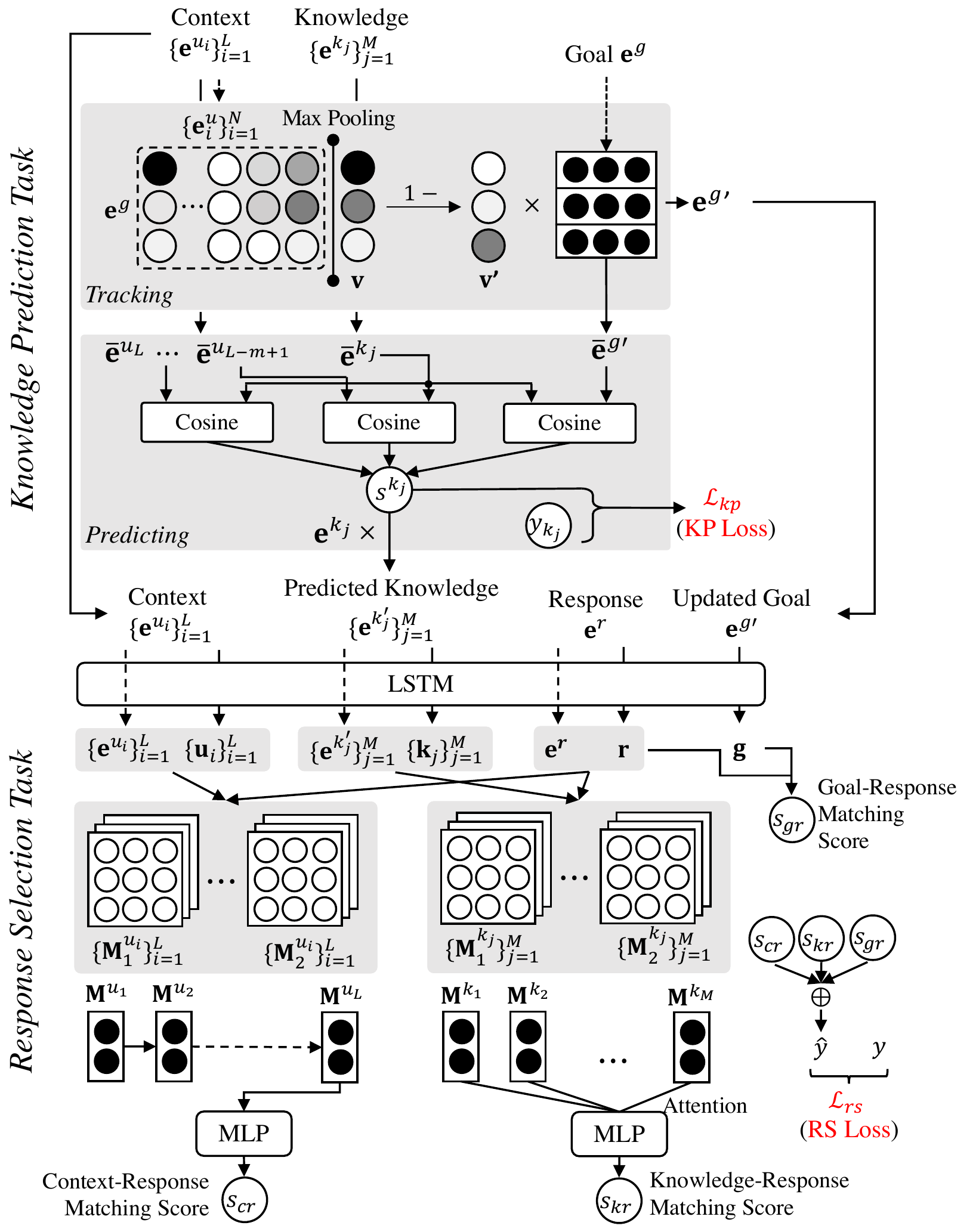}
    \setlength{\abovecaptionskip}{0.1cm}
    \setlength{\belowcaptionskip}{0cm}
    \caption{The structure of KPN. The predicted knowledge $\be^{k'_i}$ and updated goal $\be^{g'}$ in the knowledge prediction task will be used as input to the response selection task.}
    \label{fig:structure}
    \vspace{-10pt}
\end{figure}

In this work, we propose a multi-task learning framework KPN that contains \textit{response selection} (RS) and \textit{knowledge prediction} (KP) as two distinct tasks, as illustrated in Figure~\ref{fig:structure}. The predicted knowledge and updated goal from the KP task are used as input to the RS task. The loss functions in the two tasks are combined for training the model jointly. 
Different from the existing work that fuses the two tasks together and trains the whole model by only the final RS loss ($\mathcal{L}_{rs}$), we propose using a KP loss ($\mathcal{L}_{kp}$) to supervise the knowledge prediction process directly. The overall loss is as follows:
\begin{align}
    \mathcal{L} = \lambda \mathcal{L}_{kp} + \mathcal{L}_{rs},
\end{align}
where $\lambda$ is a hyperparameter (set as 0.3 in our experiment) to control the influence of the KP loss. 
The joint learning process allows us to better tell if a wrong response is obtained due to a wrong prediction of knowledge or a wrong selection of response. Details of the two tasks are presented in Sections~\ref{sec:kpt} and~\ref{sec:rs}.


The processes of KP and RS are based on the following basic representations: an utterance $u_i$ in the context, a goal $g$, a knowledge triplet $k_j$ (concatenated as a word sequence), and a response $r$ are first represented as matrices $\be^{u_i}$, $\be^{g}$, $\be^{k_j}$, and $\be^{r}$ respectively through a pre-trained embedding table. They will be used in different ways in the KP and RS processes.

\subsection{Knowledge Prediction (KP) Task}
\label{sec:kpt}
It is widely believed that knowledge can help select suitable responses.
However, not all knowledge triplets are useful in selecting responses for a conversation turn. 
Therefore, predicting whether a knowledge triplet should be used is a critical step. 

\noindent\textbf{Goal Tracking} To decide what to say in a response, one has to know what part of the goal is still uncovered. KPN achieves this by a goal tracking process (shown in the \textit{Tracking} part of Figure~\ref{fig:structure}). The basic idea is to match the goal and the context, then the mismatched entities are considered as uncovered. Concretely, we concatenate all utterances in the context as a long sequence $\{\be_i^{u}\}_{i=1}^{N}$, where $N$ is the total number of words in all the utterances, and then match it with the goal ($\be^{g}$) by cosine similarity: $\bm_{ij} = \cos(\be^{g}_i,\be^u_j)$. 
Then max-pooling is applied to extract the strongest matching signals: $v_i = \text{ReLU}( \text{Maxpooling}(\bm_{i,:}))$. The obtained values ($\bv$) represent the degree of coverage of the entities in the goal, while $\bv'=\mathbf{1}-\bv$ represents the remaining part that should be covered in the following dialogue. Finally, the vector $\bv'$ is used to update the representation of the goal: $\be^{g'}=\bv'\cdot\be^{g}$. This goal tracking method is simple but effective, and more sophisticated designs can be investigated as future work.

\noindent\textbf{Knowledge Predicting} The knowledge prediction process is shown in the \textit{Predicting} part of Figure~\ref{fig:structure}.  
The relevance of a piece of knowledge is determined by both the state of the goal and the current topic of the dialogue. The former determines the target, while the latter determines the starting point. Ideally, the relevant knowledge should pave a way leading from the current topic to the desired goal.
Usually, the current topic is contained in the last several utterances, thus we leverage them to predict the relevant knowledge.
Given the updated goal $\be^{g'}$, the last $m$ utterances
$\{\be^{u_i}\}_{i=L-m+1}^{L}$ (where $L$ is the number of utterances in the context, and $m$ is a hyperparameter set as 3 in our experiments), and the $j$-th piece of knowledge $\be^{k_j}$, we first compute their sentence-level representations by mean-pooling over word dimensions: $\bar{\be}^{u_{i}} = \text{mean}(\be^{u_{i}})$, $\bar{\be}^{g'} = \text{mean}(\be^{g'})$, and $\bar{\be}^{k_j} = \text{mean}(\be^{k_j})$. 
Then we use cosine similarity $s^{k_j}_{g'} = \cos(\bar{\be}^{g'}, \bar{\be}^{k_j})$, $s^{k_j}_{i} = \cos(\bar{\be}^{u_i}, \bar{\be}^{k_j})$ to measure their relevance,
where $i\in[L-m+1, L]$, and we obtain $(m+1)$ scores $[s^{k_j}_{g'}, s^{k_j}_{L-m+1}, \cdots, s^{k_j}_{L}]$. 
The relevance scores with respect to both the goal and the context topic are aggregated by a multi-layer perceptron (MLP) with a sigmoid activation function ($\sigma$), which is then used to update the representation of the $j$-th knowledge triplet:
\begin{align}
    s^{k_j} = \sigma\big(\text{MLP}([s^{k_j}_{0}; s^{k_j}_{L-m+1}; \cdots; s^{k_j}_{L}])\big), \quad
    \be^{k'_j} = s^{k_j}\be^{k_j},
\end{align}
where $s^{k_j}$ is the predicted probability of the $j$-th knowledge triplet to be used in the current turn. 


\noindent\textbf{Weakly Supervised Knowledge Prediction}
\label{sec:kp}
To make a correct prediction of knowledge, the common method is tuning the knowledge prediction process according to the final response selection error. The process is thus implicitly supervised~\cite{DBLP:conf/acl/LiuFCRYL18,DBLP:conf/ijcai/LianXWPW19,DBLP:conf/acl/WuGZWZLW19}. 
To further improve the learning of the knowledge prediction, besides the response selection loss, we introduce a weakly supervised knowledge prediction loss to train it explicitly. 

In practice, it is difficult to have manual labels for knowledge triplets in each dialogue turn. To address this problem, we propose a method to generate weak labels automatically. For each knowledge SPO triplet, we adopt an entity linking method to link it to the response: if the \textit{object entity} appears in the ground-truth response, we label it as $1$, otherwise as $0$\footnote{For long descriptive entities (\ie, non-factoid sentences such as the \textit{Comment} entity about \textit{Bo Peng} in the Knowledge Graph in Figure~\ref{fig:example}), if more than 70\% part is covered by the ground-truth response, we label it as one. We do not use the subject entity (\eg, ``Bo Peng''), because it is shared by many triplets, thus is less accurate as the label.}.  We assume this weak label can indicate whether such a piece of knowledge is used in the ground-truth response. 
With the weak labels $y_{k_j}$, we can compute a binary cross-entropy loss, which we call  KP loss, as follows:
\begin{align}
\small
    \mathcal{L}_{kp} = - &\frac{1}{|\mathcal{D}|} \sum \big( y_{k_j} \log s^{k_j} + (1-y_{k_j}) \log(1-s^{k_j}) \big).
\end{align}


\subsection{Response Selection (RS) Task}
\label{sec:rs}
Response selection (RS) is the main task.
As shown in Figure~\ref{fig:structure}, KPN considers the interactions between response and three types of information, \ie, the context, the knowledge, and the remaining goal. The former two can be modeled in the same way: similar to existing work~\cite{yuan-etal-2019-multi,DBLP:conf/cikm/HuaFTY020,ZhuNZDD21}, we compute matching matrices based on both the input representations ($\be^{u_i}$, $\be^{k'_j}$ and $\be^{r}$) and their sequential representations obtained by LSTM~\cite{DBLP:journals/neco/HochreiterS97}. As a result, we denote the obtained matrices as $\bM^{u}$ and $\bM^{k}$ and apply a CNN with max-pooling to extract the matching features $\bv^{u}$ and $\bv^{k}$.

(1) \textbf{Context-Response Matching}
The matching features between the context and response are aggregated by an LSTM and the corresponding final state is fed into an MLP to compute the matching score $s_{cr}$.
We use LSTM because it can model the dependency and the temporal relationship of utterances in the context. 

(2) \textbf{Knowledge-Response Matching}
Different from the context, we assume knowledge triplets to be independent. Thus, we use an attention-based method to aggregate the matching features:
\begin{align}
\small
    \alpha_i = \text{ReLU}\big(\text{MLP}(\bv^{k}_i)\big), \;\,
    \bh_2 = \sum_{i=1}^{k_M} \frac{e^{\alpha_i}}{\sum_{j}e^{\alpha_j}}\bv^{k}_i, \;\,
    s_{kr} = \text{MLP}(\bh_2).
\end{align}
This way, a knowledge triplet that is more related to the response will have a higher weight in the aggregated features and contributes more in computing the final matching score. 

(3) \textbf{Goal-Response Matching}
As the goal is a single sequence of tokens, which is much easier to model, we compute the goal-response matching score $s_{gr}$ by an MLP based on their LSTM representations at the last time step.

The final matching score is then computed as: $\hat{y} = \big(s_{cr} + s_{kr} + s_{gr}\big) / 3$. We use the binary cross-entropy loss to compute the errors:
\begin{align}
\small
    \mathcal{L}_{rs} = -\frac{1}{|\mathcal{D}|}\sum\big({y \log \sigma(\hat{y}) - \left(1-y\right) \log(1-\sigma(\hat{y}))}\big).
\end{align}

\section{Experiments}


\subsection{Datasets and Baseline Models}
We experiment on datasets DuConv and DuRecDial. \textbf{DuConv}~\cite{DBLP:conf/acl/WuGZWZLW19} is built for knowledge-grounded proactive human-machine conversation. The dialogues are about movies and stars. 
The total number of training, validation, and test samples is 898,970, 90,540, and 50,000.
\textbf{DuRecDial}~\cite{DBLP:conf/acl/LiuWNWCL20} is created as a conversational recommendation dataset, which contains dialogues between a seeker and a recommender.  
The domain of dialogue includes movie, music, food, etc. 
The number of training, validation, and test samples is 342,340, 38,060, and 55,270.
The negative responses are randomly sampled with a 1:9 positive/negative ratio in both datasets.
We compare our model against two groups of baseline methods:

\noindent\textbf{DuRetrieval}~\cite{DBLP:conf/acl/WuGZWZLW19} is the only retrieval-based model specifically designed for proactive dialogue. It uses a Transformer-based encoder for context and response representation. The conversation goal is used as an additional piece of knowledge. All knowledge triplets are represented by a bi-GRU with attention mechanism.

The other group of methods are not proposed for proactive dialogue but for general knowledge-grounded dialogue. As they also incorporate knowledge into dialogue generation, we replace our knowledge selecting module in the KP task by theirs to make a comparison. 
\textbf{MemNet}~\cite{DBLP:conf/aaai/GhazvininejadBC18} uses a memory network that performs ``read'' and ``write''  on the knowledge by matrix multiplication.
\textbf{PostKS}~\cite{DBLP:conf/ijcai/LianXWPW19} trains a knowledge prediction process to make the prior probability (using only the context) of the knowledge prediction close to the posterior probability (using both context and response).
\textbf{NKD}~\cite{DBLP:conf/acl/LiuFCRYL18} is similar to MemNet, but it first concatenates the context and knowledge representations and then uses an MLP to compute the weight for each piece of knowledge. 

\subsection{Evaluation}
All models are evaluated in two scenarios.

\noindent\textbf{On test set} Similar to the existing work~\cite{zhang-etal-2018-personalizing,DBLP:conf/acl/WuGZWZLW19}, we evaluate the performance of each model by \textbf{Hits@1}, \textbf{Hits@3}, and Mean Reciprocal Rank (\textbf{MRR}) for selecting the correct response when it is mixed up with several other candidates. Hits@$k$ measures the ratio of the ground-truth response among the top $k$ results. 

\noindent\textbf{Practical application} Following~\cite{DBLP:conf/acl/WuGZWZLW19}, we also evaluate the performance of the models in a more practical scenario, where each ground-truth utterance is mixed up with 49 utterances retrieved by Solr\footnote{\url{https://lucene.apache.org/solr/}. If the number of retrieved results is less than 49, we use random samples to pad.}. The task is to rank the ground-truth response as high as possible. This test simulates a practical scenario where the model is acting as a reranker for the candidate list returned by an upstream retrieval system.
We use several metrics to evaluate the model from different perspectives. \textbf{BLEUs} are used to measure the quality (similarity) of the response w.r.t. the ground-truth. 
To evaluate the model's ability to incorporate knowledge into dialogues, we compute the \textbf{knowledge precision/recall/F1} score used in previous studies ~\cite{DBLP:conf/ijcai/LianXWPW19,qin-etal-2019-conversing,DBLP:conf/acl/WuGZWZLW19}, which measure how much knowledge (either correct or wrong) has been used in the responses. 
We also compute a more meaningful \textbf{knowledge accuracy} to measure if the selected response uses the same piece of knowledge as that involved in the ground-truth response. 
Similarly, \textbf{goal accuracy} measures if a goal in the ground-truth is correctly covered by the selected response.


\subsection{Experimental Results}
\begin{table}[!t]
    \centering
    \small
    \setlength{\abovecaptionskip}{0cm}
    \setlength{\belowcaptionskip}{0cm}
    \caption{Evaluation results. ``KLG'' stands for knowledge, and ``Acc''  for accuracy. ``GT'' for ground-truth. ``+X'' means that the knowledge prediction module is replaced by X. The improvement obtained by KPN over DuRetr. is statistically significant with $p$-value $<0.01$ in t-test.}
    {
    \begin{tabular}{c|l|c|c|c|c|c|c}
    \hline\cline{1-8}
    {} & {} & \multirow{2}{*}{GT} & \multirow{2}{*}{DuRetr.} & \multicolumn{4}{c}{\textbf{KPN}}\\
    \cline{5-8}
    {} & {} & {} & {} & {\scriptsize Full} & {\scriptsize +MemNet} & {\scriptsize +PostKS} & {\scriptsize +NKD} \\\cline{1-8}\cline{1-8}
    \multirow{10}{*}{\rotatebox[origin=c] {90}{\footnotesize DuConv}} & \textbf{\footnotesize Hits@1} & - & 50.12 & \textbf{66.94} & 52.54 & 39.98 & 56.42 \\
    {} & \textbf{\footnotesize Hits@3} & - & 75.68 & \textbf{87.52} & 78.79 & 65.70 & 81.54 \\
    {} & \textbf{\footnotesize MRR} & - & 63.13 & \textbf{78.30} & 67.90 & 57.09 & 70.77 \\\cline{2-8}
    {} & \textbf{\footnotesize BLEU1} & 1.00 & 0.47 & \textbf{0.56} & 0.50 & 0.48 & 0.50 \\
    {} & \textbf{\footnotesize BLEU2} & 1.00 & 0.32 & \textbf{0.42} & 0.34 & 0.33 & 0.35 \\
    {} & \textbf{\footnotesize KLG. P} & 38.24 & 30.11 & \textbf{33.45} & 29.24 & 28.55 & 29.40 \\
    {} & \textbf{\footnotesize KLG. R} & 9.20 & 7.24 & \textbf{8.05} & 7.03 & 6.87 & 7.07 \\
    {} & \textbf{\footnotesize KLG. F1} & 14.83 & 11.68 & \textbf{12.97} & 11.34 & 11.07 & 11.40 \\
    {} & \textbf{\footnotesize KLG. Acc.} & 100.0 & 53.64 & \textbf{57.82} & 50.90 & 50.42 & 52.94 \\
    {} & \textbf{\footnotesize Goal Acc.} & 100.0 & 58.90 & \textbf{77.58} & 72.36 & 69.44 & 74.62 \\\cline{1-8}\cline{1-8}
    \multirow{10}{*}{\rotatebox[origin=c]{90}{\footnotesize DuRecDial}} & \textbf{\footnotesize Hits@1} & - & 77.38 & \textbf{91.50} & 75.34 & 82.45 & 82.74 \\
    {} & \textbf{\footnotesize Hits@3} & - & 89.02 & \textbf{98.86} & 93.92 & 96.60 & 97.03 \\
    {} & \textbf{\footnotesize MRR} & - & 84.07 & \textbf{95.18} & 85.00 & 89.58 & 89.96 \\\cline{2-8}
    {} & \textbf{\footnotesize BLEU1} & 1.00 & 0.46 & \textbf{0.61} & 0.51 & 0.53 & 0.53 \\
    {} & \textbf{\footnotesize BLEU2} & 1.00 & 0.39 & \textbf{0.51} & 0.39 & 0.41 & 0.41 \\
    {} & \textbf{\footnotesize KLG. P} & 52.64 & 43.42 & \textbf{52.55} & 41.04 & 43.70 & 42.87 \\
    {} & \textbf{\footnotesize KLG. R} & 3.76 & 3.10 & \textbf{3.76} & 2.93 & 3.12 & 3.07 \\
    {} & \textbf{\footnotesize KLG. F1} & 7.02 & 5.79 & \textbf{7.01} & 5.48 & 5.83 & 5.72 \\
    {} & \textbf{\footnotesize KLG. Acc.} & 100.00 & 94.90 & \textbf{95.35} & 94.32 & 94.90 & 94.81 \\
    {} & \textbf{\footnotesize Goal Acc.} & 100.00 & 78.34 & \textbf{84.96} & 82.58 & 83.12 & 83.93 \\\cline{1-8}\cline{1-8}
    \end{tabular}
    }
    \label{tab:re}
    \vspace{-10pt}
\end{table}
The evaluation results are shown in Table~\ref{tab:re}. 
Based on the results, we can observe: (1) KPN outperforms all baselines significantly by achieving the highest scores on all evaluation metrics. (2) Compared with DuRetrieval, KPN improves Hits@1, Hits@3, and MRR by a large margin. This strongly indicates that KPN has a better capability of selecting correct responses. (3) In the practical application scenario, according to the results on BLEU, we can conclude that KPN can select responses that are more similar to the golden responses. (4) On knowledge prediction, as a comparison, we also provide the evaluation result of the ground-truth. We find that our method outperforms other knowledge prediction models (MemNet, PostKS, and NKD) on knowledge P/R/F1 and accuracy. This demonstrates that the explicit supervised knowledge prediction is more effective than the implicit ones used in the other methods. Nevertheless, there is still a big gap between our results and the ground-truth, showing that the process could be much improved. 


\noindent\textbf{Reliability of the Weak Labels}
As we use an entity linking method to automatically generate weak labels for knowledge prediction, to evaluate the reliability of these labels, we randomly select 100 samples comprising 1,437 knowledge triplets from the validation set of DuConv, and ask three human annotators to label which triplet is necessary to select the current response. The result indicates that 90.26\% of the generated labels are consistent with human annotations\footnote{The Fleiss Kappa is 0.698 that indicates the annotators achieve a substantial agreement.}. This demonstrates the high reliability of the labels automatically generated by our entity linking method. 

We carried out detailed \textbf{Ablation Study} and \textbf{Influence of Hyperparameter}, showing that both the goal and knowledge strongly impact the final results. Due to space limit, these experiments
are presented in our Github page.

\section{Conclusion}
In this paper, we proposed a new approach to 
retrieval-based proactive dialogue. 
In our model, we define two tasks for response selection and knowledge prediction. An interactive matching structure is applied to model the matching between the knowledge and the response. 
In order to make a good prediction of knowledge, explicit supervision signals are used, which are derived from the ground-truth responses.
Experimental results demonstrated that our model can achieve better performance than the baseline models in which the two tasks are mixed up. In particular, it is shown that training the knowledge prediction explicitly is very effective. This work is a first demonstration of the importance of modeling knowledge and goals explicitly in proactive dialogue.


\begin{acks}
We thank Wenquan Wu and Zhen Guo for the insightful suggestions. 
This work was supported by a Discovery grant of the Natural Science and Engineering Research Council of Canada, National Natural Science Foundation of China (No. 61872370 and No. 61832017),  Beijing Outstanding Young Scientist Program (NO. BJJWZYJH012019100020098), and Shandong Provincial Natural Science Foundation (Grant ZR2019ZD06).
\end{acks}

\balance

\end{document}